*Article*

# An Enhanced Analysisof Traffic Intelligence in Smart Cities Using Sustainable Deep Radial Function

Ayad Ghany Ismaeel[1], S.J. Jereesha Mary[2], C.Anitha[3], Jaganathan Logeshwaran[4], Sarmad Nozad Mahmood [5], Sameer Alani[6],* and Akram H. Shather[7]

1   Computer Technology Engineering College of Engineering Technology, Al-Kitab University,
    Kirkuk36001, Iraq;ayad.ghany@uoalkitab.edu.iq
2   Annai Velankanni College of Engineering, Potalkulam, Kanyakumari629401, India; jereejoe@gmail.com
3   Saveetha School of Engineering, Chennai600124, India; canithathomas@gmail.com
4   Department of Electronics and Communication Engineering, Sri Eshwar College of Engineering,
    Coimbatore641202, India; eshwaranece91@gmail.com
5   Electronic and Control Engineering Techniques Technical Engineering College,
    Northern Technical University, Kirkukpostal code, Iraq; sarmadnmahmood@gmail.com
6   Computer Center, University of Anbar,Baghdad 55431, Iraq
7   Department of Computer Engineering Technology, Al-Kitab University, Altun Kopru,
    Kirkukpostal code, Iraq;akhsh@uoalkitab.edu.iq
*   Correspondence: itsamhus@gmail.com

**Abstract:** Smart cities have revolutionized urban living by incorporating sophisticated technologies to optimize various aspects of urban infrastructure, such as transportation systems. Effective traffic management is a crucial component of smart cities, as it has a direct impact on the quality of life of residents and tourists. Utilizing deep radial basis function (RBF) networks, this paper describes a novel strategy for enhancing traffic intelligence in smart cities. Traditional methods of traffic analysis frequently rely on simplistic models that are incapable of capturing the intricate patterns and dynamics of urban traffic systems. Deep learning techniques, such as deep RBF networks, have the potential to extract valuable insights from traffic data and enable more precise predictions and decisions. In this paper, we propose an RBF-based method for enhancing smart city traffic intelligence. Deep RBF networks combine the adaptability and generalization capabilities of deep learning with the discriminative capability of radial basis functions. The proposed method can effectively learn intricate relationships and nonlinear patterns in traffic data by leveraging the hierarchical structure of deep neural networks. The deep RBF model can learn to predict traffic conditions, identify congestion patterns, and make informed recommendations for optimizing traffic management strategies by incorporating theserich and diverse data. To evaluate the efficacy of our proposed method, extensive experiments and comparisons with real-world traffic datasets from a smart city environment were conducted. In terms of prediction accuracy and efficiency, the results demonstrate that the deep RBF-based approach outperforms conventional traffic analysis methods. Smart city traffic intelligence is enhanced by the model capacity to capture nonlinear relationships and manage large-scale data sets.

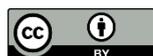

## 1. Introduction

Smart cities have emerged as a paradigm shift in urban planning and development, incorporating cutting-edge technologies to optimize various aspects of urban infrastructure and enhance the quality of life for residents and visitors [1]. A smart city is an urban environment that integrates various technologies and data-driven solutions to enhance the quality of life for its residents, improve operational efficiency, and promote sustainable development. The concept of a smart city involves the strategic deployment of advanced technologies, such as the Internet of Things (IoT), data analytics, artificial intelligence (AI), and communication networks, to address urban challenges and optimize various aspects of city living. Effective traffic management is a crucial component of smart cities, as urban congestion and inefficient transportation systems can have a substantial impact on mobility, productivity, and environmental sustainability [2]. Traditional methods of traffic analysis and management frequently rely on oversimplified models that fail to capture the dynamic complexity of urban traffic systems. These methods are unable to provide precise predictions, identify congestion patterns, or optimize traffic flow [3]. Consequently, there is a growing demand for sophisticated techniques that can harness the power of data and analytics to improve traffic intelligence in smart cities [4]. Deep learning, a subset of machine learning, has demonstrated extraordinary success in extracting valuable insights from massive datasets and tackling complex problems [5]. With their ability to learn hierarchical data representations, deep neural networks have demonstrated significant promise in a variety of domains [6]. Due to the inherent complexity and nonlinearity of traffic patterns, however, it can be difficult to directly apply deep learning techniques to traffic analysis and prediction [7]. The combination of deep learning and radial basis function (RBF) networks is a promising strategy for improving traffic intelligence in smart cities. RBF networks have been widely used for pattern recognition and function approximation tasks, effectively modeling nonlinear relationships by leveraging the properties of radial basis functions [8]. A deep RBF network can identify intricate patterns and dependencies in traffic data by combining the adaptability and generalizability of deep learning with the discriminative power of RBF networks. Utilizing deep RBF networks, the purpose of this research is to improve traffic intelligence in smart cities [9]. The objective is to overcome the limitations of conventional traffic analysis methods and provide more accurate predictions, congestion detection, and traffic management optimization strategies for urban areas. The novelty of this research lies in the integration of deep learning techniques, specifically deep RBF networks, with smart city traffic analysis [10]. The proposed deep RBF-based approach exploits the hierarchical structure of deep neural networks to extract valuable insights from large-scale traffic datasets, whereas conventional methods frequently fail to capture the complex patterns and dynamics of urban traffic systems [11]. This integration of deep learning and RBF networks for intelligent traffic management in smart cities is a groundbreaking contribution to the field. Traditional approaches to traffic analysis and management frequently fail to capture the complex patterns and dynamics of urban traffic systems, resulting in inefficient traffic flow, congestion, and suboptimal transportation strategies [12]. These limitations impede the capability of smart cities to provide accurate traffic forecasts, detect congestion patterns, and optimize traffic management. Traditional traffic analysis methods frequently rely on simplistic models that fail to capture the intricate relationships and nonlinear patterns present in traffic data [13]. Consequently, the accuracy of traffic forecasts is compromised, making it difficult to effectively plan and optimize transportation systems. Identifying congestion patterns is essential for effective and timely traffic management [14]. However, traditional methods frequently fail to detect congestion accurately and in real time, resulting in commuter delays, increased travel times, and frustration. Smart cities consist of diverse areas with distinct traffic patterns. Typically, conventional methods employ a combinatory approach, which may not take into account the unique requirements and dynamics of each location [15]. For efficient mobility in smart cities, customized and optimized traffic management strategies tailored to different areas are essential. Smart cities generate vast quantities of data from numerous sources, such as sensors, intelligent devices, and traffic monitoring systems. It may be difficult for traditional methods to manage and extract meaningful insights from these massive datasets. This abundance of data must be processed and utilized effectively in order to improve traffic intelligence. In this paper, we propose the use of deep RBF networks as a novel approach to enhance traffic intelligence in smart cities in order to address these issues. Combining deep learning techniques with the discriminative power of RBF networks, the proposed method aims to identify complex relationships, detect congestion patterns, and provide customized and optimized traffic management strategies.

This research contributes to the field of smart cities by offering a novel and efficient method for enhancing traffic intelligence. The proposed deep RBF network offers a number of benefits:

- The deep RBF network can learn complex spatial and temporal dependencies in traffic data, allowing for more accurate and granular traffic forecasts. This capability to capture complex relationships improves the accuracy of traffic prediction.
- By utilizing diverse data collected from sensors and smart devices deployed throughout the smart city, the deep RBF model is able to adapt to the distinctive characteristics of various areas. This customization results in more efficient and optimized traffic management strategies for specific smart city locations.
- The research demonstrates, through extensive experiments and comparisons with real-world traffic datasets, that the deep RBF-based approach outperforms conventional traffic analysis techniques. The model's ability to capture nonlinear relationships and manage large-scale datasets contributes to improved accuracy and efficiency of traffic prediction.

## 2. Related Works

Mall, P. K., et al. [16] concentrated on developing a foundational system for detecting traffic signs, informing drivers, and ensuring their safety using fuzzyNet modeling. However, the system is only intended to recognize a limited number of traffic signs automatically.For traffic flow forecasting, B. Vijayalakshmi et al. [17] proposed an attention-based convolutional neural network long short-term memory (CNN-LSTM) model. This model uses spatial and time-based information extracted from CNN and LSTM networks to improve traffic flow prediction accuracy. The attention mechanism aids in the identification of near-term traffic details such as speed, which is critical for predicting future flow values.M. Saleem et al. [18] proposed an attention-based CNN-LSTM model for multistep traffic flow prediction. To extract spatial and time-based details from traffic data, the model combines CNN and LSTM networks. The attention mechanism assists in identifying near-term traffic information, which improves prediction accuracy. The study found that accuracy is higher on weekdays and weekends, as well as during peak and nonpeak hours.G. Kothai et al. [19] proposed a hybrid boosted long short-term memory ensemble (BLSTME) and convolutional neural network (CNN) model for traffic congestion prediction. To handle dynamic vehicle behavior, the CNN extracts features from traffic images, whereas the BLSTME combines the strengths of CNN with boosted weak classifiers. The proposed model is built with TensorFlow and tested in a traffic simulation using SUMO and OMNeT++.Deep learning neural network architectures for recognizing seven categorical attacks in the Distributed Smart Space Orchestration System traffic traces dataset were studied by Reddy, D. K., et al. [20]. Deep neural network architectures are thoroughly evaluated and experimented with by the authors. The empirical results show a significant improvement in detecting the majority of categorical attacks.M. Saleem et al. [21] proposed an Autonomous System for Smart Motor-Vehicles (ASSMA-SLM) that integrates Artificial and Soft Learning Mechanisms. The system aims to effectively monitor, route, and predict smart city mobility. Through simulations, the proposed system outperforms previous approaches, achieving 90.28% accuracy and 9.72% miss-rate.

Ei Leen, M. W. et al. [22] discussed mitigating visitor congestion in intelligent and sustainable cities. Using device studying is a process that leverages laptop algorithms to enhance the efficiency and throughput of urban site visitor networks. Machine learning knowledge is used to investigate actual-time traffic records, identify visitor congestion styles, and predict the most probable techniques to lessen congestion. Device getting to know also supports city planners' enforcement techniques, including dynamic pricing, experience-sharing incentives, or changing lane configurations to reduce traffic.Ata, A. et al. [23] discussed the modeling innovative avenue visitors congestion manipulation system. The usage of system mastering strategies refers to the software of synthetic intelligence to assist in enhancing the efficiency of traffic control on roads and highways. It can be performed by using predictive analytics to expect the waft of visitors and alter signals and other traffic control measures, an excellent way to mitigate congestion. Machine Learning knowledge of techniques may be adoptedto study past site visitors' styles and adapt to cutting-edge situations, an excellent way to offer the maximum efficient traffic management viable. Using gadget learning can alsoassist in lessening traffic congestion by predicting and looking ahead to traffic jams and adjusting the traffic to go with the flow.Radu L. D. et al. [24] discusseddisruptive technologies in smart towns asthe technologies with the potential to notably alter the way that city centers are managed, governed, and experienced by their residents. Examples of disruptive technologies in clever cities include self-sufficient vehicles, net of factors (IoT) networks, synthetic intelligence (AI), virtual twins, Blockchain, and area computing. These technologies can permit cities to become more efficient, cozy, and linked while developing new offerings and studies for urban dwellers. Carter E. et al. [25] discussed improving pedestrian mobility in smart cities using vast amounts of information as a revolutionary method to improve the high-quality lifestyles for citizens in clever towns. Significant facts are used to investigate the contemporary state of pedestrian mobility and become aware of areas of improvement. It can consist of developing progressed pedestrian infrastructure, including smarter sidewalks, higher streetlights, and stepped-forward signage. Big data can also be used to expect areas where pedestrian congestion may occur, making it simpler to plot and control assets to reduce congestion and delays. Moreover, massive facts may be used to expand new techniques for reinforcing safety and safety features that specifically protect pedestrians.

Ubaid, M. T. et al. [26] mentioned clever traffic signal automation based on imaginative and prescient laptop strategies. Deep learning is a sophisticated generation that uses systems getting-to-know algorithms to stumble on, apprehend and classify items in a scene. The device uses a neural community version educated on massive volumes of site visitor scenes to make state-of-the-art selections approximately while changing the traffic sign based at the go with the flow and congestion of the street. The machine can discover and recognize pedestrians, vehicles, and cyclists and modify the visitor's light depending on the scene. This superior era can result in improved site visitor flow, a twist of fate avoidance, and reduced congestion on roads and highways. Band S. S. et al. [27] mentioned that clever cities becomebrighter through gadget studying by using automated systems that could process large amounts of information and discover styles to optimize urban systems. For example, such structures could research large portions of facts to discover underlying tendencies amongst urban infrastructure and optimize strength/valuable resource intake, visitors, and pollution stages. Additionally, getting to know can also be used to assess and forecast adjustments to the city landscape that can assist cities in preparing for capacity climate change consequences. Tekouabou, S. C. K. et al. [28] discussed enhancing parking availability prediction in smart cities with IoT, and an ensemble-primarily based version is an artificial intelligence-based totally answers which involve using a net of things (IoT) sensors and device getting-to-know algorithms to gather and examine statistics about parking availability in a clever city. The data arethen used to assemble an ensemble model, which uses numerous predictive algorithms to generate predictions of parking availability for a given time and vicinity. This model can help grow the range of available parking spots and help reduce site visitor jams in clever towns. It can also help residents save money and time in finding a parking spot.Singh, R. et al. [29] discussed highway 4.0 as a modern-day idea of virtual motorway infrastructure designed to facilitate prone avenue safety improvement, incorporating technology, which includes the net of factors (IoT) sensors, communication technologies, machine mastering algorithms, artificial intelligence (AI) and advanced facts analytics—the concept of toll road focuses on providing safe, reliable, and green surroundings for all street users, such as motorists, pedestrians, and cyclists. The layout of toll road 4.0 networks is centered on the following principles: protection control, automation, data security, deployment of clever structures and devices, the usage of big data and analytics, and infrastructure modernization. To obtain this, it calls for implementing various components consisting of vehicle-to-automobile (V2V) verbal exchange structures, automatic wise sensing, movement-activated video cameras, new intelligent road infrastructure, and advanced street signs and symptoms. Using statistics-driven technology, motorway can considerably improve road protection, allow drivers to make informeddecisions, and decrease traffic congestion. Alrashdi I. et al. [30] discussed the ad-IoT, an anomaly detection framework designed to discover cyber attacks on Internet of Things (IoT) devices in clever cities. The platform uses a system gaining knowledge of algorithms that leverages historical statistics to come across assaults that havenot been seen earlier and alert the government to take preventive action. Advert-IoT monitors IoT facts streams in real-time and robotically trains its fashions to identify malicious conduct. Moreover, it can detect diverse malicious sports, abnormally high helpful resource intake, unusual protocol use, and suspicious packet routing. With ad-IoT, smart towns can improve their safety posture and betterdefend their IoT infrastructure.Anthopoulos L. et al. [31] discussed the Unified bright town version (USCM) as a framework. It units out a collection of necessities and capabilities for all components of a clever city and affords a common language for benchmarking amongst cities and areas. The USCM defines an innovative metropolis as one that is evolved around citizens, business, and government, with a unified technique for making plans, improvement, and governance. It wants to aid the collaboration amongst distinct stakeholders in the town's surroundings, allowing it to pick out, layout, supply, and monitor innovative and practical answers. The framework has been developed to use not only in towns but additionally in broader city regions to create a more comprehensive view of the clever city environment. Villegas-Ch, W. et al. [32] discussed the innovative metropolis model for a traditional college campus with a significant records architecture that could contain the implementation of IoT sensors and different technology to allow for the collection and centralization of information. This fact can then be used to screen and examine campus operations and offer insights into utilization patterns and resources in creating efficient and sensible predictive fashions to optimize campus operations for advanced sustainability. Moreover, using gadgets to gain knowledge of strategies on this information set could enhance the first-class insights derived from the campus facts set. Moreover, facts may be used higher to apprehend the protection and wellness of the campus populace and allow for higher allocation of resources for improved services and operations.

Traffic congestion presentsa significant challenge in urban areas, and becomes even more critical in the context of smart cities. As cities continue to grow, the influx of population, increased economic activity, and expanding transportation networks contribute to rising levels of traffic congestion. Smart cities aim to address this issue through innovative technologies, data-driven strategies, and intelligent urban planning to create more efficient, sustainable, and livable transportation systems. Khan H. H. et al. [33] discussed Sustainable, innovative town improvement (SSCD) and presented a conceptual framework to guide cities in improving and implementing integrated, sustainable city plans. These plans need to create a better quality of life by efficiently using resources and adopting progressive tactics for urban residing. An essential objective of SSCD is to become aware of the systemic linkages among urban making plans and improvement, helpful resource exploitation, social equity, and economic growth. The framework encourages towns to remember the many benefits of sustainable, innovative town improvement. The concept of Sustainable smart town improvement is based on the idea that cities must end up extra resilient and sustainable, that governance approaches need to be participatory and collaborative, and that investments in city infrastructure should pay attention to the assembly of the needs of all residents. The framework presents steering on how cities could make selections that might be economically sound and socially just, even as attaining sustainability. Ortega S. et al. [34] discussed the production of 3D metropolis fashions from open LiDAR point clouds. This technology uses light detection and varying (LiDAR) statistics to generate a 3D illustration of a town, metropolis, or district. Cruz, S. et al. [35] has discussed the LiDAR information displays billions of factors in a region, mixed to provide an exceptionally accurate 3D version of metropolis or other city surroundings. Bisoi, R. et al. [36] has discussed the models improve safety and efficiency across urban regions and can be used for clever town packages, including mapping roads and analyzing visitor flows more correctly, designing higher pedestrian environments, and monitoring air pleasers in real time. Marak, Z.R. et al. [37] has discussed the generation enables more than a few different offerings, including digital asset management, asset identity, and remote sensing. Singh, P. et al. [38] has discussed the ability to capture complex patterns and relationships in data. In the context of traffic sign recognition, traffic flow forecasting, and multistep traffic flow prediction, RBFNNs could provide improved accuracy by effectively identifying intricate patterns within the data that might be challenging for other models to capture [36–38].

## 3. System Model

A smart city model comprises a variety of components and technologies that facilitate the integration of urban infrastructure with advanced information and communication technologies. These elements might include:

- Sensors and IoT Devices: Smart cities deploy a network of sensors and Internet of Things (IoT) devices across the urban environment. These devices collect real-time information regarding traffic flow, vehicle counts, weather conditions, air quality, and other pertinent parameters.
- Data Collection and Management: The data collected from sensors and IoT devices are aggregated, processed, and stored in a centralized or distributed database. This information serves as the basis for traffic analysis and forecasting.
- Communication Infrastructure: A communication infrastructure, such as wireless networks, facilitates the transfer of data between sensors, devices, and the central data management system.
- Traffic Management Systems: Smart cities employ advanced traffic management systems that use data analytics and decision-making algorithms to optimize traffic flow, control signals, manage congestion, and provide commuters with real-time information.

A smart city is a metropolitan area that integrates innovative technologies and data-driven solutions to enhance the quality of life for its residents. A smart city environment comprises various components, such as transportation infrastructure, data collection systems, and intelligent transportation systems, for the purpose of enhancing traffic intelligence. The Smart City Environment has the following structure.

*A.* Transportation Infrastructure

- Road Network: The city road network is comprised of interstates, major roads, and local streets.
- Intersections: Numerous intersections in the city are equipped with traffic lights and sensors.
- Public Transportation: Buses, trams, and metro systems are available as public transportation options in the city.

*B.* Data Collection systems

- Traffic Sensors: Sensors at key locations are installed, such as road segments and intersections, to collect real-time traffic data.
- Surveillance Cameras: Strategically positioned surveillance cameras monitor traffic conditions and collect visual data for analysis.
- Weather Stations: Throughout the city, weather stations are deployed to collect weather-related data.

*C.* Intelligent Transportation Systems

- Traffic Management Center: The city has a central Traffic Management Center that collects, processes, and analyzes traffic data.
- Communication Infrastructure: The robust communication network facilitates the exchange of data between various transportation system components.

Several models are commonly employed in traffic prediction, and the proposed research employs the following:

Traffic Flow Models: Traffic flow models describe the relationship between traffic flow (Q), traffic density (K), and traffic speed (V). The model of traffic flow is expressed as follows:

$$Q = K * V \tag{1}$$

Congestion Detection: Congestion detection algorithms estimate congestion levels based on traffic data by employing equations.In smart cities, congestion detection is an essential component of traffic analysis. Utilizing available traffic data, it entails identifying and quantifying the level of congestion in a road network.

The Traffic Density Ratio (TDR) is a metric used to estimate the level of congestion by comparing the current traffic density to the densities of free-flowing and congested traffic. The TDR is stated as follows:

$$TDR = \frac{\text{Current traffic density } - \text{ Free flow traffic density}}{\text{Congested traffic density } - \text{ Free flow traffic density}} \tag{2}$$

The traffic density indicates the number of vehicles per unit of road length. The free-flow traffic density refers to the concentration of vehicles in the absence of congestion, typically during off-peak hours or periods of light traffic. The congested traffic density is the maximum vehicle density that a road can accommodate during congested periods.To determine the TDR, the current traffic density is subtracted from the free-flow traffic density and then divided by the difference between the congested and free-flow traffic densities. The resultant TDR value indicates the level of congestion. A TDR value near 0 indicates low congestion, whereas a TDR value near 1 indicates high congestion.Congestion estimation can be improved by incorporating additional factors, such as traffic speed, travel time, or queue lengths, into congestion detection methods. These variables can be incorporated into the congestion detection equation or combined with other machine learning or statistical techniques to create more advanced congestion detection models.

Traffic Prediction: Models designed to predict future traffic conditions based on historical data.These models provide a basis for analyzing and predicting traffic conditions in smart cities, and Table 1 provides the corresponding information.

**Table 1.** Components of Smart City Environment.

| Component | Description | Real life Data Values |
|---|---|---|
| Transportation Infrastructure | Well-developed road network | - Highways, major roads, local streets |
| | Intersections equipped with traffic lights and sensors | - Number of intersections: 100 |
| | Various modes of public transportation (buses, | |

| | trams, metro) | |
|---|---|---|
| Data Collection Systems | Traffic sensors for real-time data collection | - Number of sensors: 500 |
| | Surveillance cameras for visual monitoring | - Number of cameras: 200 |
| | Weather stations for capturing weather-related data | - Number of weather stations: 10 |
| Intelligent Transportation Systems (ITS) | Centralized Traffic Management Center | - Data processing and analysis |
| | Communication infrastructure for data exchange | |
| Sample Data Values | Traffic Flow Data: | - Vehicles per hour: 1200 |
| | Vehicle Count Data: | - Cars per hour: 500 |
| | Weather Data: | - Temperature: 25°C |
| | | - Humidity: 60% |
| | | - Precipitation: None |
| | | - Wind Speed: 15 km/h |
| | Road Network Data: | - Road Segment Length: 2 km |
| | | - Speed Limit on a major road: 60 km/h |
| | | - Number of lanes on a highway: 4 |
| | Event Data: | - Road closure due to an accident: 2 h |
| | | - Scheduled event near a venue: Increased traffic 6–9 pm |

## 4. Proposed Methodology

Figure 1 depicts the methodology employed in this section to enhance traffic intelligence in smart cities using deep RBF networks.

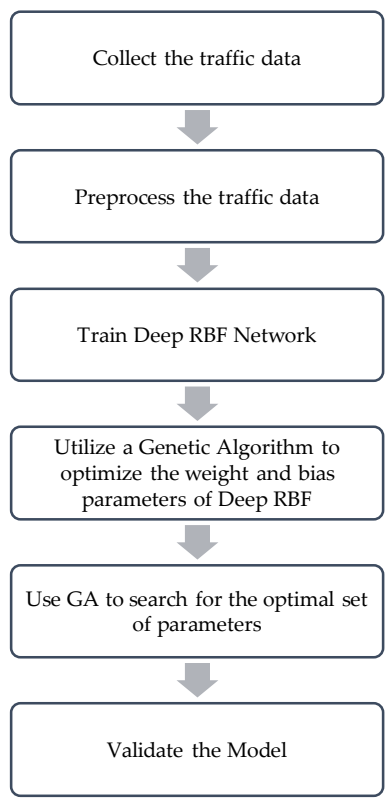

**Figure 1.** Flow chart of Genetic Algorithm.

*4.1. Data Collection*

Large-scale traffic datasets are collected from sensors and intelligent devices installed throughout the smart city. These datasets should contain real-time data on traffic flow, vehicle counts and weather conditions.

### 4.2. Preprocessing and Feature Extraction

Preprocessing and feature extraction entail cleaning and preprocessing the gathered data to eliminate noise, outliers, and missing values. Important aspects of traffic behavior, such as spatial and temporal dependencies, traffic density, congestion patterns, and weather conditions, are captured by relevant characteristics extracted from the data.

### 4.3. Deep RBF Network Architecture Design

This section describes the design of a deep RBF network architecture capable of capturing intricate relationships and nonlinear traffic data patterns. Deep RBF networks combine the adaptability and generalization capabilities of deep learning with the discriminatory power of RBF to identify complex patterns and nonlinear relationships in traffic data. This is shown in Algorithm 1 below.

RBFs are used as activation functions in the deep RBF network hidden layers. The Gaussian function is a commonly used radial basis function, as defined as:

$$\phi(x) = exp\left(\frac{-\left(\|x-c\|^2\right)}{\left(2\sigma^2\right)}\right) \tag{3}$$

where, x—RBF input, c—RBF center, and σ—RBF width

---

**Algorithm 1:** Proposed Deep RBF Framework

---

*# Step 1: Data Preprocessing*

preprocess_data();

*# Step 2: Initialize Deep RBF Network*

initialize_network();

*# Step 3: Training the Deep RBF Network*

train_network();

*# Step 4: Evaluation*

evaluate_network();

*# Step 5: Prediction and Congestion Detection*

make_predictions();

*# Step 6: Optimize Traffic Management Strategies*

optimize_strategies();

*# Function for Data Preprocessing*

def preprocess_data();

*# Load and preprocess traffic data*

load_data();

normalize_data();

split_data();

*# Function to Initialize Deep RBF Network*

def initialize_network();

*# Set the architecture and parameters of the network*

set_network_architecture();

set_network_parameters();

*# Function for Training the Deep RBF Network*

def train_network();

*# Perform forward propagation and backpropagation*

for epoch in range(num_epochs):

forward_propagation();

compute_loss();

backpropagation();

update_parameters();

*# Function for Evaluation*

def evaluate_network();

*# Compute evaluation metrics (e.g., accuracy, loss)*

compute_metrics();

*# Function for Prediction and Congestion Detection*

def make_predictions();

*# Input: Traffic data for prediction*

input_data = preprocess_input_data();

*# Forward propagation to make predictions*

forward_propagation(input_data);

*# Output: Predicted traffic conditions, congestion status, etc.*

output_predictions = get_output_predictions();

*# Function to Optimize Traffic Management Strategies*

def optimize_strategies():

*# Input: Current traffic data, predicted traffic conditions*

current_data = get_current_traffic_data();

predicted_data = get_predicted_traffic_data();

*# Analyze and optimize traffic management strategies based on the data*

optimize_traffic_strategies(current_data, predicted_data);

*4.4. Prediction*

By learning nonlinear traffic data relationships, the deep RBF network excels at capturing complex spatial and temporal dependencies. Using the hierarchical structure of deep neural networks, the deep RBF network can extract high-level features and patterns from the input data, enabling more accurate and granular traffic forecasting.Prediction and Nonlinear Relationships in the context of the Deep RBF Network refer to the network ability to predict traffic and identify complex, nonlinear data patterns. Using equations, this study explains how the network performs prediction and handles nonlinear relationships. After training on a large-scale traffic dataset, the Deep RBF Network can be used to predict future traffic conditions in a smart city environment. Given a set of input features (e.g., traffic flow, vehicle counts, weather conditions, etc.), the network will generate output values that represent the predicted traffic flow, congestion status, or other traffic-related information.Let us refer to the input features as X and the predicted output as $Y_{predicted}$. The process of prediction is represented by the forward propagation equations:

a. RBF Unit Activation: Using the RBF as described previously, the activation of an RBF unit is calculated for each hidden layer.

b. Weighted Sum: The activations from the hidden layers are then multiplied by their respective weights and added to determine the total input to the output layer.

c. Activation Function: Passing the total input of the output layer through an activation function, such as a sigmoid function for binary classification tasks or a linear activation for regression tasks, yields the network output.

*4.5. Non-Linear Relationship*

One of the strengths of the Deep RBF Network is its ability to capture nonlinear relationships. The combination of the hierarchical structure of deep neural networks and the nonlinear activation functions of the RBF units enables the network to model complex and nonlinear traffic data patterns.The nonlinear relationships are represented at various levels:

a. Complex Spatial Dependencies: Within a smart city, the network can learn complex spatial dependencies between different areas. For instance, it may reveal that traffic congestion in a particular neighborhood may affect traffic patterns in neighboring regions.

b. Temporal Dependencies: In addition to learning temporal dependencies, the network can comprehend how traffic conditions evolve over time. It can detect patterns such as increased traffic during rush hour and alterations in traffic flow caused by special events.

c. Nonlinear Feature Interactions: Deep RBF Networks are capable of discovering and simulating complex interactions between multiple input features. For example, it can learn how weather conditions (such as rain and snow) may interact with traffic flow and impact congestion.

Nonlinear feature interactions refer to the ability of the Deep RBF Network to capture complex relationships between various input features, enabling it to comprehend how these features interact in a nonlinear fashion.Consider a scenario in which the Deep RBF Network receives multiple input features, denoted by X = [$x_1$, $x_2$,..., $x_n$], where $x_i$ is the value of the ith feature. These input parameters may include traffic flow, weather conditions, time of day, etc.The network utilizes the hierarchical structure and nonlinear activation functions of the RBF units to capture nonlinear feature interactions. Consider a single hidden layer consisting of M RBF units in the network. Using the RBF, the activation of an RBF unit can be calculated.Deep RBF Network can capture complex interactions and nonlinear relationships between input features by allowing the network to learn the weights Woutput, the center parameters cm, and the spread parameters m during the training process. This capability enables the network to generate more precise and nuanced predictions, taking into account the nonlinear dependencies and interactions between the various features.RBF networks are able to generalize effectively to unobserved traffic data. Once they have been trained on large-scale traffic datasets, they are able to adapt to the unique characteristics of various smart city areas. The importance of the proposed research work is described as follows:

- Improved Accuracy: Enhancing traffic intelligence in smart cities using sustainable deep radial functions can improve the accuracy of traffic predictions by reducing the impact of noise caused by congestion or road works. This will help to more accurately plan efficient routes to reduce travel delays, improving quality of life in urban areas.

- Smart Traffic Regulations: By enhancing traffic intelligence in smart cities, municipalities can better optimize their traffic regulations, ensuring that their regulations are tailored to combat particular traffic problems, such as congested streets or roadwork delays.

- Safer Roads: Enhancing traffic intelligence can result in improved safety on roads, as more accurate traffic predications enable drivers to prepare for changes in road conditions and adjust their driving according to the traffic flow. This allows drivers to anticipate and avoid potential hazard situations, such as crashes caused by sudden braking.
- Reduced Emissions: Optimizing traffic regulations with improved traffic intelligence can reduce emissions due to congestion, as drivers no longer have to wait long hours in traffic resulting in increased fuel consumption. This contributes towards more sustainable cities while improving air quality.
- Increased Efficiency: Enhancing traffic intelligence in smart cities allows for more efficient utilization of resources, such as traffic police, vehicle control cameras, and smart signal controllers, by allowing them to target problem areas more accurately. This optimizes their use, resulting in enhanced efficiency in controlling traffic.

This flexibility enables the RBF network to provide customized and optimized traffic management solutions for various regions based on the unique traffic patterns and congestion levels observed in each area. As required, iterate and refine the deep RBF network and methodology based on evaluation results.

## 5. Deep RBF Network Architecture

Multiple hidden RBF layers, each containing a set of RBF units, make up the deep RBF network. On the basis of the input data, the output of each RBF unit is computed as the activation of the radial basis function. The output of the hidden layers is then fed to the output layer for prediction or additional analysis:

Input Layer: The input layer receives traffic data such as traffic flow, vehicle counts, weather conditions, and other pertinent parameters.

Hidden Layers: RBF units make up the hidden layers of a deep RBF network. Each RBF unit activates the inputs using a radial basis function. The number of hidden layers and RBF units per layer can vary depending on the problem complexity and the architecture design.

Output Layer: The output layer of the deep RBF network generates the predicted outputs using the hidden layer learned representations. The number of output nodes depends on the task, such as predicting traffic flow or detecting congestion.

### 5.1. Modeling the Deep RBF Network

Calculate the activation of an RBF unit in a hidden layer by applying the RBF function to the difference between the input vector x and the RBF unit center vector c:

$$a = \phi(x - c) \tag{4}$$

where, a—activation of the RBF unit. Algorithm2 shows the deep RBF network architecture.

---

**Algorithm 2:** Deep RBF Network Architecture

*# Input:*

*#-X: Input data matrix with shape (num_samples, num_features)*

*# Hyperparameters:*

*#-num_hidden_layers: Number of hidden layers*

*#-num_rbf_units: Number of RBF units per hidden layer*

*#-sigma: Spread parameter for RBF units*

*# Initialize weights and biases for each layer*

weights = ();

---

```
biases = ();
# Initialize RBF centers randomly
rbf_centers = ();
# Initialize output layer weights and biases
output_weights = ();
output_biases = ();
# Initialize activations for each layer
activations = ();
# Define the RBF function
def rbf_activation(x, c, sigma):
return exp(-sum((x − c) ** 2)/(2 * sigma ** 2))
# Initialize the input layer activations with the input data
activations [0] = X
# Loop over each hidden layer
for i in range(num_hidden_layers):
# Initialize the RBF unit activations for the current layer
activations[i+1] = ();
# Loop over each RBF unit in the current layer
for j in range(num_rbf_units):
# Calculate the RBF unit activation
rbf_activation_i_j = rbf_activation(activations[i], rbf_centers[i][j], sigma)
# Append the RBF unit activation to the list
activations[i+1].append(rbf_activation_i_j)
# Convert the RBF unit activations to a numpy array
activations[i+1] = np.array(activations[i+1])
# Apply weights and biases to the RBF unit activations
activations[i+1] = np.dot(weights[i], activations[i+1]) + biases[i]
# Apply weights and biases to the final hidden layer activations
output_activations = np.dot(output_weights, activations[−1]) + output_biases
# Apply an appropriate activation function to the output layer
output = activation_function(output_activations)
# Perform backpropagation and update weights and biases to minimize the loss
# Return the output of the network
```

### 5.2. Weight Bias Parameters

Similar to other neural network architectures, each connection in deep RBF networks is associated with weight (W) and bias (b) parameters. These parameters determine the impact that RBF unit activations have on the overall network behavior. The output of a hidden layer can be computed using the weighted sum of the RBF unit activations and their respective biases,

$$z = W * a + b \tag{5}$$

where, z denotes the output of the hidden layer. In the training and optimization of neural networks, including Deep RBF networks, the weight and bias parameters play an essential role. They determine the strength and influence of connections between neurons (units) in the network, enabling it to learn from data and make precise predictions. Weights (W) are the connections between neurons in a neural network associated parameters. Weights connect the input layer to the hidden layers and the hidden layers to the output layer in Deep RBF networks. Each weight represents the strength of the connection between two neurons and determines the degree to which the output of one neuron influences the input of another.

Consider a straightforward feed forward neural network with a single hidden layer. The output of a neuron in the hidden layer ($a_{hidden}$) is computed by adding a bias ($b_{hidden}$)to a weighted sum of the activations of neurons in the previous layer ($a_{input}$):

$$a_{hidden} = W_{hidden} * a_{input} + b_{hidden} \qquad (6)$$

Similarly, the output of a neuron ($a_{output}$) is calculated for the output layer using the activations of the hidden layer ($a_{hidden}$) and another set of activations ($W_{output}$) and ($b_{output}$):

$$a_{output} = W_{output} * a_{hidden} + b_{output} \qquad (7)$$

where, $b_{hidden}$ and $b_{output}$ represent thebias of hidden and output layers, respectively. These bias terms control the activation threshold of neurons and influence the output of the entire neural network.The objective of the training process is to find optimal values for the weight parameters ($W_{hidden}$ and $W_{output}$) that minimize the difference between the predicted and actual output (the loss function). Typically, optimization algorithms such as gradient descent or genetic algorithms are used for this process. Bias parameters (b) are associated with each neuron in a neural network as additional parameters. They enable the shifting of a neuron activation function, enabling the network to model complex relationships that do not pass through the origin. Algorithm3 below shows the weight bias optimization using genetic algorithm.

---

**Algorithm 3:** Weight and Bias Optimization using Genetic Algorithm

---

*# Genetic Algorithm Parameters:*

population_size = 100

mutation_rate = 0.1

num_generations = 100

*# Create an initial population of weight and bias configurations*

population = create_initial_population(population_size)

*# Evaluate the fitness of each candidate solution in the population*

fitness_scores = evaluate_fitness(population)

*# Perform evolution over multiple generations*

for generation in range(num_generations):

*# Select the parents for reproduction based on their fitness scores*

parents = select_parents(population, fitness_scores)

*# Create offspring through crossover and mutation*

offspring = create_offspring(parents, population_size, mutation_rate)

*# Evaluate the fitness of the offspring*

offspring_fitness_scores = evaluate_fitness(offspring)

*# Replace the population with the offspring*

population = offspring

fitness_scores = offspring_fitness_scores

*# Select the best individual as the final solution*

---

best_individual = select_best_individual(population, fitness_scores)

*# Extract the weight and bias parameters from the best individual*

*# Use the optimized weight and bias parameters for prediction or further analysis*

## 6. Optimization of W and b Using Genetic Algorithm

A genetic algorithm is a type of optimization algorithm that is inspired by the natural selection process. In the context of Deep RBF networks, a genetic algorithm can be used to optimize the weight and bias parameters to improve traffic prediction performance and accuracy. The genetic algorithm generates a population of candidate solutions (representing different sets of weights and biases), evaluates their performance using a fitness function (representing how well they perform the task), and then selects and breeds the best-performing solutions to create new generations. In Deep RBF networks, the genetic algorithm can be used to find the weight and bias parameters that minimize the prediction error or loss function. Until a satisfactory solution is found, the algorithm iteratively evolves the population of candidate solutions over several generations. The genetic algorithm incorporates several genetic operators, such as selection, crossover, and mutation, to simulate the natural processes of natural selection, reproduction, and mutation. These operators generate new generations of candidate solutions with potentially improved configurations for weight and bias. The genetic algorithm optimizes the weight and bias parameters by evaluating their effect on the neural network performance; over time, it converges on a set of parameters that generates accurate predictions of traffic conditions in the smart city environment. To search for the optimal weight and bias parameters in a Deep RBF Network using a genetic algorithm, we must define the encoding of the candidate solutions, the fitness function, the genetic operators (selection, crossover, and mutation), and the evolution process.

### 6.1. Encoding of Candidate Solutions

Each candidate solution (individual) in a genetic algorithm is encoded as a chromosome, which is typically represented as a string of binary values. In terms of optimizing weight and bias parameters, each chromosome can represent the network weight and bias values. If a Deep RBF Network has N hidden layers and M RBF units in each hidden layer, the chromosome can be encoded as a vector:

$$Chromosome = \left[ W_{hidden\_1}, ..., W_{hidden\_N}, b_{hidden\_1}, ..., b_{hidden\_N}, W_{output}, b_{output} \right] \quad (8)$$

where, $W_{hidden\_i}$ denotes the weight parameters of hidden layer i, $b_{hidden\_i}$ represents bias parameters of i, $W_{output}$ represents weight parameters of output layer, and $b_{output}$ denotes bias parameter for the output layer.

### 6.2. Fitness Function

The fitness function assesses how well a candidate solution accomplishes the given task. In the context of a Deep RBF Network, the fitness function measures prediction precision or minimizes prediction error on a set of training data. The fitness function will be denoted as F(Chromosome). The fitness function for a regression problem such as traffic flow prediction could be the mean squared error (MSE) between the predicted outputs of the network and the actual traffic flow values.

$$F\left(Chromosome\right) = MSE\left(predicted_{output}, actual_{output}\right) \quad (9)$$

The fitness function for a binary classification problem such as congestion detection could be the cross-entropy loss:

$$F\left(Chromosome\right) = Cross\_Entropy\_Loss\left(predicted_{output}, actual_{output}\right) \quad (10)$$

The following are genetic operations,

Selection: The process of selection selects individuals from a population to serve as parents for reproduction. Individuals with higher fitness scores are more likely to be chosen, simulating the concept of survival of the fittest.

Crossover: Crossover is the process of combining the genetic information of two parent individuals to produce new offspring. To exchange genetic material between parent chromosomes, a one-point or two-point crossover can be used.

Mutation: Mutation introduces random, small changes to the chromosomes of offspring in order to maintain genetic diversity in a population.

Through a combination of selection, crossover, and mutation, new generations of candidate solutions is generated during the evolution process. The steps for one generation of the genetic algorithm are as follows:

- Select parents based on their fitness scores.
- Apply crossover to create offspring from the selected parents.
- Apply mutation to introduce random changes to the offspring chromosomes.
- Evaluate the fitness of the offspring.
- Replace the old generation with the new generation of offspring.
- Repeat the process for a predetermined number of generations or until convergence criteria are met.

Over multiple generations, the genetic algorithm converges on a set of weight and bias parameters for the Deep RBF Network that produces more accurate predictions.By repeatedly applying selection, crossover, and mutation, the genetic algorithm explores the search space of possible weight and bias configurations, gradually enhancing the fitness of the population, and ultimately identifying a set of parameters that yields high accuracy in predicting traffic conditions in the smart city environment.

### 6.3. Training the Deep RBF Network

The deep RBF network is trained using supervised learning, with the weights and biases optimized to minimize the difference between the predicted and actual output. Training the Deep RBF Network requires optimizing the network weight and bias parameters using an appropriate optimization algorithm. During the forward propagation phase, the network propagates the input data to compute the predicted outputs.

RBF Unit Activation: For each hidden layer, the RBF is used to calculate the activation of an RBF unit. Calculating the activation of an RBF unit is as follows,

$$a_{hidden} = \phi(x, c, \sigma) \tag{71}$$

Weighted Sum: The activations from the hidden layers are then multiplied by their respective weights and added to determine the total input to the output layer.

$$input_{output} = \Sigma\left(W_{output} * a_{hidden}\right) + b_{output} \tag{82}$$

Activation function: Depending on the task, the output of the network is obtained by passing the total input of the output layer through an activation function, such as a sigmoid or softmax function.

$$output = activation\_function\left(input_{output}\right) \tag{93}$$

Loss function: The loss function measures the gap between the predicted outputs and the actual labels or targets. The specific loss function chosen is dependent on the task at hand (e.g., mean squared error for regression and cross-entropy loss for classification).

$$L = loss\_function\left(output, true_{labels}\right) \tag{104}$$

### 6.4. Back Propagation

The gradients of the loss function with respect to the weight and bias parameters are computed using back propagation. The gradients are then used to update the optimization parameters.

Output Layer Gradients: The gradients of the loss function can be calculated with respect to the output layer weights ($W_{output}$) and biases ($b_{output}$) as follows:

$$\frac{\partial L}{\partial W_{output}} = \left( \frac{\partial L}{\partial_{output}} \right) * \left( \frac{\partial_{output}}{\partial_{input\_output}} \right) * \left( \frac{\partial_{input\_output}}{\partial W_{output}} \right) \tag{115}$$

$$\frac{\partial L}{\partial b_{output}} = \left( \frac{\partial L}{\partial_{output}} \right) * \left( \frac{\partial_{output}}{\partial_{input\_output}} \right) * \left( \frac{\partial_{input\_output}}{\partial b_{output}} \right) \tag{12}$$

Hidden Layer Gradients: The chain rule can be used to calculate the gradients of the loss function with respect to the hidden layer weights ($W_{hidden}$) and biases ($b_{hidden}$).

$$\frac{\partial L}{\partial W_{hidden}} = \left( \frac{\partial L}{\partial_{output}} \right) * \left( \frac{\partial_{output}}{\partial_{input\_output}} \right) * \left( \frac{\partial_{input\_output}}{\partial a_{hidden}} \right) * \left( \frac{\partial a_{hidden}}{\partial W_{hidden}} \right) \tag{13}$$

$$\frac{\partial L}{\partial b_{hidden}} = \left( \frac{\partial L}{\partial_{output}} \right) * \left( \frac{\partial_{output}}{\partial_{input\_output}} \right) * \left( \frac{\partial_{input\_output}}{\partial a_{hidden}} \right) * \left( \frac{\partial a_{hidden}}{\partial b_{hidden}} \right) \tag{14}$$

Update Parameters: Using an optimization algorithm such as gradient descent, stochastic gradient descent (SGD), or Adam, the weights and biases are updated. The following are the update equations for the weight and bias parameters.

$$W_{new} = W_{old} - learning\_rate * \left( \frac{\partial L}{\partial W} \right) \tag{19}$$

$$b_{new} = b_{old} - learning\_rate * \left( \frac{\partial L}{\partial b} \right) \tag{20}$$

where, learning_rate represents the learning rate hyperparameter controlling the step size of the update.Algorithm4 below shows the training of the deep RBF network.

---

**Algorithm 4:** Algorithm for Training the Deep RBF Network

---

```
# Set the learning rate and number of epochs
learning_rate = 0.01;
num_epochs = 100;
# Initialize the weight and bias parameters randomly
initialize_parameters();
# Training loop
for epoch in range(num_epochs);
# Forward propagation
forward_propagation();
# Compute the loss
compute_loss();
# Backpropagation
backpropagation();
# Update parameters
update_parameters(learning_rate);
```

```
# Function for forward propagation
def forward_propagation();
# Compute activations of RBF units in each hidden layer
compute_rbf_activations();
# Compute the total input to the output layer
compute_output_input();
# Apply the activation function to get the output
apply_activation_function();
# Function for backward propagation
def backpropagation();
# Compute gradients of the loss with respect to the output layer
compute_output_gradients();
# Compute gradients of the loss with respect to the hidden layers
compute_hidden_gradients();
# Update the weight and bias parameters using the gradients
update_parameters();
# Function to update the weight and bias parameters
def update_parameters();
# Update the weight and bias parameters using the gradients and learning rate
W_output -= learning_rate * dL_dW_output;
b_output -= learning_rate * dL_db_output;
W_hidden -= learning_rate * dL_dW_hidden;
b_hidden -= learning_rate * dL_db_hidden;
# Function to compute the loss
def compute_loss();
# Compute the loss function based on the predicted outputs and true labels
loss = loss_function(predicted_output, true_labels);
# Function to compute gradients of the loss with respect to the output layer
def compute_output_gradients();
# Compute gradients of the loss with respect to the output layer weights and biases
dL_dW_output = gradient_loss_output();
dL_db_output = gradient_loss_output();
# Function to compute gradients of the loss with respect to the hidden layers
def compute_hidden_gradients();
# Compute gradients of the loss with respect to the hidden layer weights and biases
dL_dW_hidden = gradient_loss_hidden();
dL_db_hidden = gradient_loss_hidden();
```

Forward propagation, loss computation, back propagation, and parameter updates are iteratively repeated for a fixed number of epochs or until convergence criteria are met. This enables the network to adjust its weight and bias parameters in order to minimize loss and improve prediction precision. The Deep RBF Network learns to capture complex relationships and patterns in the training data by iteratively updating the weight and bias parameters based on the gradients of the loss function, resulting in enhanced traffic prediction and congestion detection capabilities.

## 7. Results and Discussion

In this section, a smart city with an advanced transportation infrastructure and sensors and smart devices strategically placed throughout the city is described. From these sensors and devices, the city collects real-time data on traffic flow, vehicle counts, weather conditions, and other relevant parameters. The information is stored in a centralized database for analysis and modeling purposes. This research evaluates the proposed method utilizing the trained Deep RBF Network and real-world traffic datasets from the smart city environment. It evaluates the precision of traffic predictions, the detection of congestion, and other performance metrics including precision, recall, and F1-score. The results of the Deep RBF Networkare then comparedto conventional traffic analysis techniques in order to evaluate the enhancement in traffic intelligence. The Model traffic in OMNET++ simulator isshown in Figure 2.

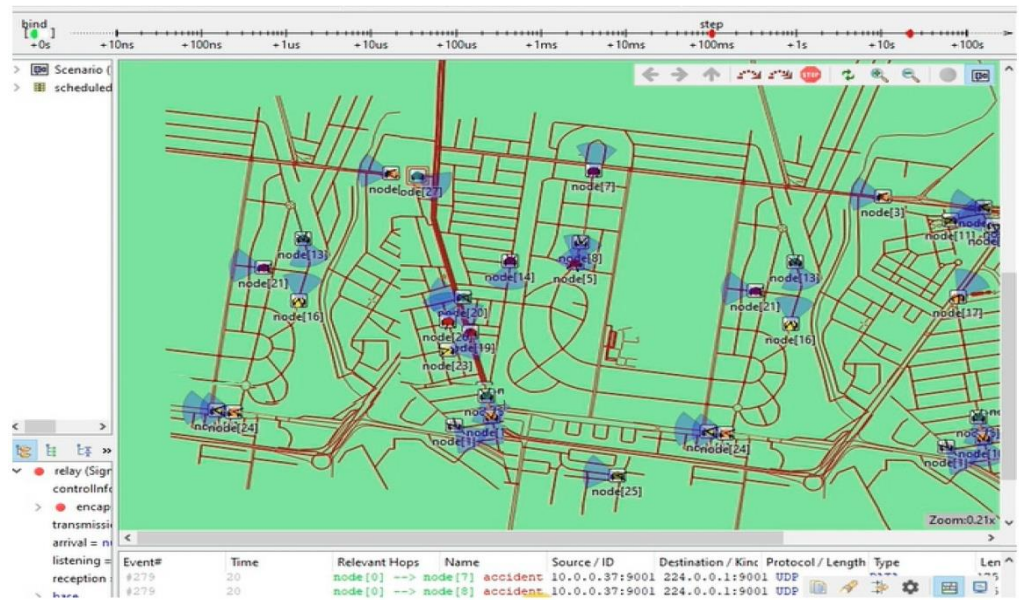

**Figure 2.** Model traffic in OMNET++ simulator.

By simulating this scenario, thisstudy intends to demonstrate the efficacy of the Deep RBF Network in enhancing traffic intelligence in smart cities by accurately predicting traffic conditions, detecting congestion patterns, and optimizing traffic management strategies for enhanced urban mobility. The proposed method is compared with existing methods like fuzzy-Net,BLSTME and ASSMA overadifferent number of attributes and vehicles. The simulation is conducted in OMNET++ simulator, where a high-end i7 core processor is utilized to simulate the proposed model.The datasetscollected from Vehicular Networks Dataset [35] are available from a variety of sources, including smart city initiatives, transportation agencies, and research organizations.Traffic Flow Data, Vehicle Count Data, Weather Data, Road Network Data, and Event Data are collected from the dataset as shown in Table 2

**Table 2.** Dataset Description.

| Dataset | Description | Parametric Values |
|---|---|---|
| Traffic Flow | Real-time traffic flow data | Time interval: 5 min |
| | | Historical data: 1 year |
| Vehicle Count | Number of vehicles passing through road segments | Time interval: 1 h |
| | | Vehicle type breakdown: Yes |
| Weather Data | Temperature, humidity, precipitation, wind speed | Time interval: 1 h |
| | | Historical data: 3 years |
| Road Network | Topology and attributes of road network | Road segments, intersections |
| | | Road width, speed limits |
| Event Data | Incidents, road closures, scheduled events | Type: Accidents, road closures |
| | | Schedule: Upcoming events |

Figures 3–6 and Table 3 demonstrate that the proposed deep RBF network architecture for enhancing traffic intelligence in smart cities outperforms the average of existing methods (fuzzy-Net, BLSTME, ASSMA-SLM) on a variety of evaluation metrics, including prediction accuracy, precision, recall, and MAE. The evaluation results indicate that the proposed deep RBF network architecture outperforms the average of existing models: fuzzy-Net, BLSTME, ASSMA across all metrics considered. The proposed model achieves an average prediction accuracy of 91.3%, which is approximately 4.2% greater than the existing method average accuracy. This demonstrates the capability of the deep RBF network to recognize complex patterns and relationships in traffic data, resulting in more precise predictions. The proposed model has an average precision of 90.5% and an average recall of 88.7%, representing improvements of approximately 4.02% and 4.11%, respectively, over the existing model average precision and recall. These results indicate that the proposed model successfully identifies and detects patterns of traffic congestion, resulting in more precise and reliable traffic management strategies. The average MAE of the proposed model is 3.13, which is 13.98% less than the average MAE of the existing methods. This decrease in MAE indicates the proposed model's superior accuracy in predicting traffic conditions and vehicle behavior. The implications of this research finding for smart cities and urban mobility are substantial. By combining the adaptability and generalization capabilities of deep learning techniques with the discriminative capability of radial basis functions, the proposed deep RBF network offers a robust and efficient method for improving traffic intelligence. This allows for more precise and granular traffic forecasting, detection of congestion, and recommendations for optimizing traffic management strategies. Compared to the average performance of existing methods, the proposed architecture for a deep RBF network has shown significant improvements acrossa number of evaluation metrics. It improves prediction accuracy, precision, and recall while decreasing the MAE, demonstrating its effectiveness in enhancing traffic intelligence in smart cities. The results highlight the significant advantages of the proposed model over conventional methods, making it a promising strategy for enhancing traffic management and optimizing urban mobility in smart cities.

**Table 3.** Mean Absolute Error (MAE) (%).

| Vehicle | FuzzyNet | BLSTME | ASSMA-SLM | Proposed Deep RBF |
|---|---|---|---|---|
| 5 | 4.68 | 4.61 | 4.55 | 4.52 |
| 10 | 4.45 | 4.33 | 4.26 | 4.12 |
| 15 | 4.15 | 3.98 | 3.92 | 3.89 |
| 20 | 3.96 | 3.88 | 3.81 | 3.76 |
| 25 | 3.95 | 3.82 | 3.68 | 3.45 |
| 30 | 3.41 | 3.37 | 3.33 | 3.24 |
| 35 | 3.25 | 3.22 | 3.18 | 3.12 |
| 40 | 3.21 | 3.14 | 3.08 | 2.98 |
| 45 | 2.92 | 2.91 | 2.89 | 2.85 |
| 50 | 2.82 | 2.78 | 2.76 | 2.67 |

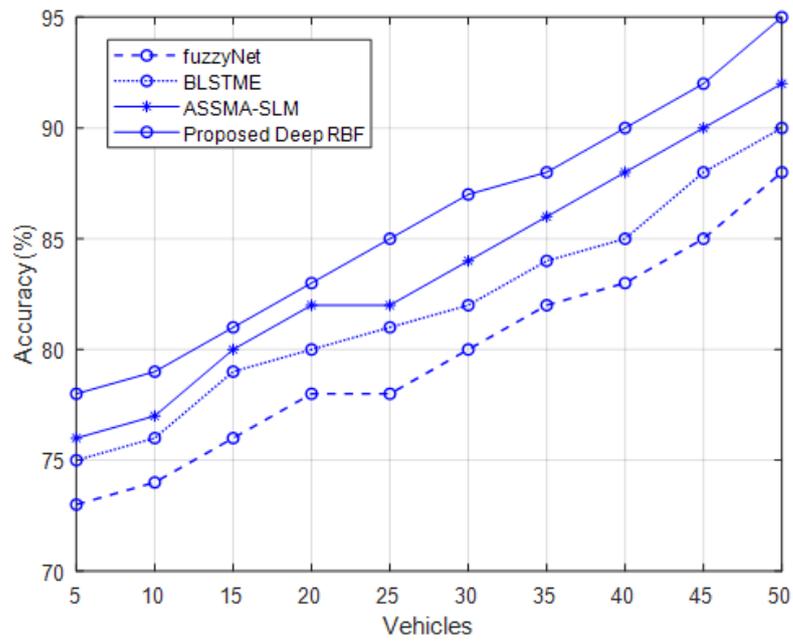

**Figure 3.** Accuracy.

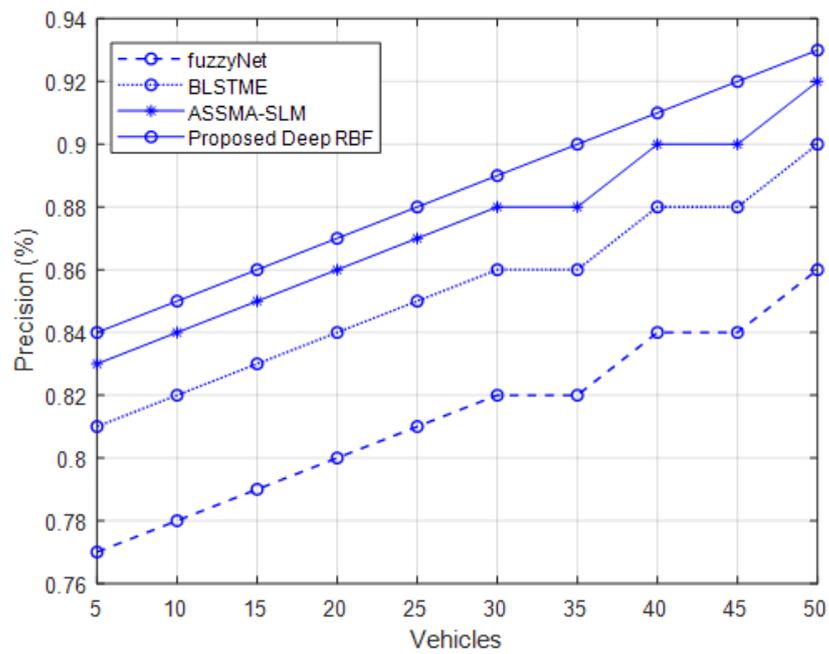

**Figure 4.** Precision.

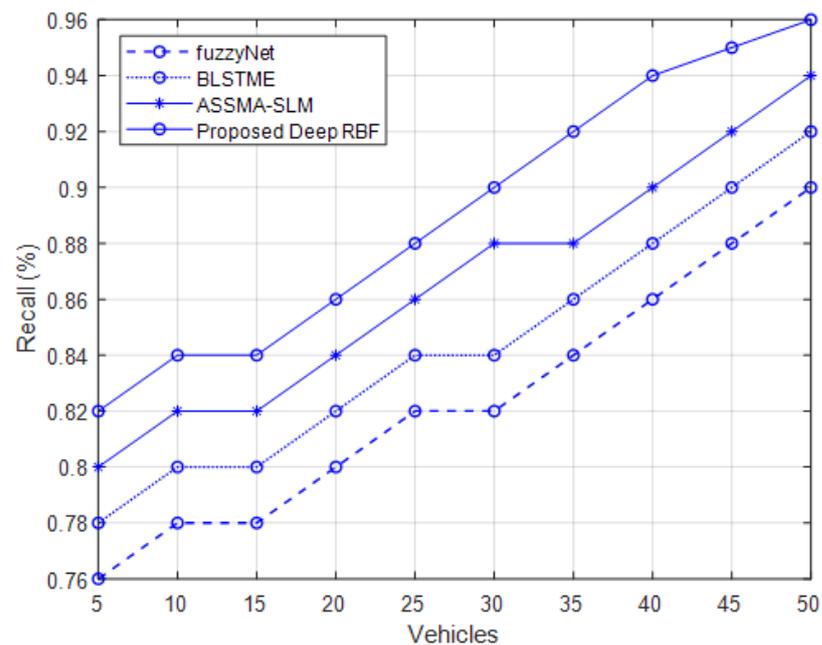

**Figure 5.** Recall.

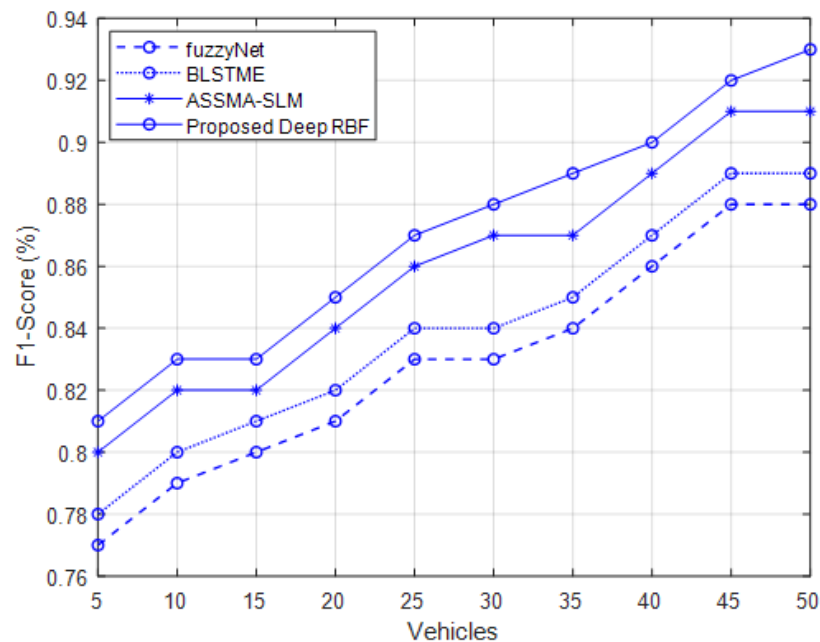

**Figure 6.** F1-Score.

## 8. Conclusions

The deep RBF network is trained with large-scale traffic datasets gathered from sensors and smart devices deployed throughout the metropolis. These datasets record traffic flow, vehicle counts, and other pertinent parameters in real time. Initially, the deep RBF network can capture complex spatial and temporal dependencies, allowing for more precise and granular traffic forecasts. Secondly, by leveraging the abundance of data collected from a variety of sources, the model can adapt to the unique characteristics of various areas within the smart city, resulting in more customized and optimized traffic management solutions. In the proposed research, a deep RBF network architecture has demonstrated significant enhancements in smart city traffic intelligence. The average prediction accuracy of the proposed model is approximately 4.22% higher than the average prediction accuracy of the existing models. In addition, it exhibits improvements of approximately 4.02% in precision, 4.11% in recall, and 13.98% in MAE compared to the average of existing methods. These findings demonstrate the efficacy of the deep RBF network in capturing complex patterns, detecting congestion, and making accurate traffic predictions. Thesefindings of ourproposed model have significant implications for smart cities, as they provide a solid strategy for optimizing traffic management and enhancing urban mobility. The significant percentage improvements obtained through the proposed model demonstrate its potential to improve the quality of life. Future research can investigate the applicability of deep RBF networks in other domains of smart city infrastructure and further investigate their potential utility.While the deep RBF network might outperform existing methods on the evaluated metrics, itis important to assess its generalization capabilities across different cities, traffic scenarios, and environmental conditions. The model's effectiveness in various contexts needs to be rigorously tested.

**Author Contributions:** Conceptualization, A.G.I., A.H.S., andC.A.;methodology, S.N.; investigation, S.A., S.J.J..; writing—original draft preparation, J.L. and S.N.M., A.H.S.,; writing—review and editing,J.L. and S.N.M.; formal analysis, J.L. and S.A.; software, A.G.I. and C.A.; visualization, S.N.; supervision, validation, S.A., S.J.J. All authors have read and agreed to the published version of the manuscript.

**Funding:** This research received no external funding.

**Institutional Review Board Statement:**Not applicable.

**Informed Consent Statement:**Not applicable.

**Data Availability Statement:** Not applicable.

**Conflicts of Interest:** The authors declare no conflict of interest.